\documentclass[conference]{IEEEtran}
\IEEEoverridecommandlockouts
\usepackage{cite}
\usepackage{amsmath,amssymb,amsfonts}
\usepackage{algorithm}
\usepackage{algpseudocode}
\usepackage{graphicx}
\usepackage{subcaption}
\usepackage{caption}
\usepackage{textcomp}
\usepackage{xcolor}
\usepackage{float}
\usepackage{parskip}

\def\BibTeX{{\rm B\kern-.05em{\sc i\kern-.025em b}\kern-.08em
    T\kern-.1667em\lower.7ex\hbox{E}\kern-.125emX}}
\begin{document}

\title{A novel Reservoir Architecture\\for Periodic Time Series Prediction}

\author{Zhongju Yuan\textsuperscript{1}, Geraint Wiggins\textsuperscript{2, 3 *}, Dick Botteldooren\textsuperscript{1*}\thanks{*Corresponding author: Geraint Wiggins, Dick Botteldooren.}\\
  \textsuperscript{1}WAVES Research Group, Ghent University, Gent, Belgium\\
  \textsuperscript{2}AI Lab, Vrije Universiteit Brussel, Belgium\\
  \textsuperscript{3}EECS, Queen Mary University of London, UK\\
  \texttt{zhongju.yuan@ugent.be},
\texttt{geraint.wiggins@vub.be},
\texttt{dick.botteldooren@ugent.be}\\
}

\maketitle

\begin{abstract}
This paper introduces a novel approach to predicting periodic time series using reservoir computing. The model is tailored to deliver precise forecasts of rhythms, a crucial aspect for tasks such as generating musical rhythm. Leveraging reservoir computing, our proposed method is ultimately oriented towards predicting human perception of rhythm. Our network accurately predicts rhythmic signals within the human frequency perception range. The model architecture incorporates primary and intermediate neurons tasked with capturing and transmitting rhythmic information. Two parameter matrices, denoted as $c$ and $k$, regulate the reservoir's overall dynamics. We propose a loss function to adapt $c$ post-training and introduce a dynamic selection (DS) mechanism that adjusts $k$ to focus on areas with outstanding contributions. Experimental results on a diverse test set showcase accurate predictions, further improved through real-time tuning of the reservoir via $c$ and $k$. Comparative assessments highlight its superior performance compared to conventional models.
\end{abstract}


\section{Introduction}
\label{section:intro}

The quasi-periodical time series prediction problem involves forecasting future values of a time series that exhibits periodic patterns or cycles~\cite{brown2004smoothing} over a certain time interval. The sounds that humans use for commutation are temporally structured sequences of events, such as musical notes. Here, {\it rhythm} means the pattern of timing and stress in the amplitude envelope of an acoustic sequence~\cite{large2015neural}, which has a basic beat, the {\it tactus}, in a specific frequency range. The {\it meter} is a perceived temporal structure that includes the tactus frequency~\cite{london2012hearing}. 

This problem is of significant importance in various domains such as industrial processes and music generators~\cite{katz2016evolution}. In this paper, our primary focus lies in the prediction of uni-variate periodical time series. In humans and embodied systems, the act of striking a rhythm has to begin prior to the real beats to account for delays in the system. Hence, for synchronization with others, precise prediction is crucial. Humans are capable of rhythm and rhythm synchronization within a limited range of beat and meter frequencies, a characteristic that we want the architecture to simulate.

While some traditional machine learning and statistical models are capable of periodic time series prediction~\cite{khashei2011novel}, they often lack the necessary generalization and may require retraining for data with different frequencies~\cite{sapankevych2009time}. Currently, certain deep learning models can also perform this task, but they demand substantial computational resources for training~\cite{cai2020traffic}. Moreover, due to their excessive parameter count, their application on real-time hardware is constrained. Hence, we opt for reservoir computing~\cite{verstraeten2007experimental,maass2002real} to address this issue. In reservoir computing, reservoir weights are typically not trained. This significantly enhances training efficiency. Moreover, as the reservoir determines the overall dynamics, we postulate that it could be trained to exhibit human-like beat and meter prediction capabilities. Additionally, due to the relatively lower parameter count compared to deep learning models, it is more feasible for practical hardware implementation~\cite{9303448}.

Conventionally, reservoirs rely on randomly generated neuron connection weights~\cite{lukovsevivcius2012practical}, making them challenging to adjust and study their properties. They struggle to excel in tasks involving the prediction of periodic time series, such as rhythms. For small signals, when the non-linear activation function responds linearly, the reservoir can be approximated by a linear time-invariant system. For random connection weights, its poles are randomly distributed within a portion of the unit circle in the Z-domain and therefore not particularly tuned to represent periodic signals. 
It is desirable to control the locations of poles and thus oscillatory behavior more precisely, thereby enhancing the likelihood of obtaining predictions through weighted summation. To this end, we design the network's weight matrix inspired by a finite difference time domain discretization of wave equations~\cite{botteldooren1995finite,kowalczyk2007line}.

In our novel reservoir structure, neurons are classified into two types: primary and intermediate. The primary neurons correspond to pressure in the inspirational physical system, while the intermediate neurons map to particle velocities. To modulate information transfer, we introduce two types of weight, $c$ and $k$. In the equivalent physical system, they represent propagation speed and decay rate, respectively. This physical inspiration imposes certain constraints on the reservoir weight matrix, yet it also enables easy adjustments to the overall reservoir speed and activation level. Furthermore, distinct regions within the reservoir can be designated distinct roles (e.g., oscillation frequencies), allowing for targeted manipulation of their dynamics.

The biological plausibility of the proposed architecture can also be examined. The overall topology-preserving structure can also be observed in the human auditory cortex, where specific areas exhibit greater responsiveness to sound modulated at various frequencies~\cite{janata2002cortical,schonwiesner2009spectro}.  At the level of individual biological neurons, the intermediate artificial neurons can be considered as synapses with a temporal storage capacity. This temporary information storage capability enables the intermediate neurons to model delays, a crucial aspect of biological neural networks~\cite{madadi2018dendritic}. The response of the primary neurons is further modulated by a global parameter that may be analogous to neurotransmitter control. This combination of structures enables the generation of slow rhythms without compromising the biological constraints of individual biological neurons. Additionally, the double-parameter architecture can discriminate between low and high spontaneous activity neurons, which exhibit varying sensitivities~\cite{taberner2005response}.

The output weights undergo training using a constructed dataset containing plausible rhythms across diverse music styles. Following the training phase, although the beat and meter may be well predicted, synchronization to the future signal may fail. The proposed reservoir structure has the advantage that we can rely on the physical equivalent to easily tune its parameters to speed up or slow down. To this end, we propose a new loss function. In both `too early' and `too late' cases, we accumulate errors using the slopes of both the prediction and target. Ultimately, by comparing the magnitudes of these two losses, we adjust $c$ and thus slow down or speed up until synchronization is reached. In the proposed reservoir structure, specific topographic areas strongly respond to specific beat frequencies. Areas that are useful for modeling typical musical rhythms can thus be promoted. Adjusting the synaptic transmission strength (denoted as $k$)  within the intermediate neurons surrounding these areas facilitates this process. We propose a dynamic selection (DS) mechanism, which selects oscillators within the reservoir and modulates the transmission strength of neighboring neurons, thereby refining the predictive efficacy of the model.

We conducted experiments on a fixed test set consisting of various beat-frequency time series. Our model achieved accurate predictions of these rhythms within a certain frequency range that matches human capabilities. Furthermore, we obtained even more precise results through the proposed adjustments to $c$ and $k$. To demonstrate the effectiveness of the proposed reservoir architecture, we compared our model to a randomly generated reservoir and a model utilizing a motor generator based on differential equations~\cite{egger2020neural}. Our model outperformed the other algorithms. We also conducted studies on the impact of $c$ and $k$ on the results, demonstrating their effectiveness in both parts.

In summary, our main contributions are as follows:
\begin{itemize}
    \item A novel reservoir structure, inspired by physics, that addresses periodic time series prediction tasks.
    \item A novel way to synchronize the prediction with the target based on a single parameter overall weight tuning in the reservoir and a new loss function.
    \item A dynamic selection (DS) mechanism to direct the reservoir's focus towards areas that play a more significant role in accurate prediction.
\end{itemize}

\section{Related work}
\label{section:related work}

The prediction of periodic time series is of paramount significance and finds widespread applications in industries \cite{hamzaccebi2009comparison}. This type of problem holds particular relevance in rhythm perception. Humans tend to respond to the occurrence of events to synchronize their actions and sounds \cite{matthews2022perceived}, such as learning to tap along with the beat of a drum. However, movement preparation is often slower than the rate of musical events, necessitating the prediction of future events for behavior preparation \cite{palmer2022we}. Commonly, there are two approaches to performing this task: predictive coding (PC) and dynamical systems (DS).

PC is a theoretical framework in neuroscience that posits that the brain constantly generates predictions about sensory inputs based on prior knowledge, with the error resulting from the disparity between expectations and predictions. The lower area of the brain receives input signals from the sensory system, while the higher level processes the projections. During this process, the error is computed \cite{vuust2018now}, serving as fundamental information for tuning periodic behavior \cite{koelsch2019predictive}. Some studies have utilized the inhibition of human actions when they should not occur, while enhancing actions when they should, as a means to engage in PC \cite{koelsch2019predictive}. A study conducted by Heggli, Konvalinka, et al. \cite{heggli2019musical} investigated dyadic partners' internal models during synchronized tapping of identical periodic patterns. Partners were exposed to either matching or mismatching metrical contexts, influencing their top-down predictions, and resulting in small mean negative asynchronies. Another study by Elliott et al. \cite{elliott2014moving} supported a Bayesian causal inference model with four free parameters, indicating that participants aimed to minimize asynchrony variance and extract a rhythmic pattern.

DS elucidates how models evolve over time, encompassing the analysis of physical synchrony \cite{pikovsky2001universal} and biological synchrony in rhythms and motor coordination \cite{ egger2020neural}. As the oscillators are coupled and facile in sharing information with one another, synchronization is also anticipated even in the absence of external inputs \cite{tal2017neural}. Non-linearity allows the oscillators to synchronize at higher resonances with input rhythmic patterns, such as 1:2 and 1:3, compared to the 1:1 synchronization in a linear system \cite{large2008resonating}. In Roman et al.'s research \cite{roman2019delayed}, individuals synchronized with a metronome, utilizing a single Hopf oscillator with a time delay to emulate unidirectional coupling. This model successfully emulates varying levels of predictive ability between musicians and non-musicians across different metronome rates.

The DS theory emphasizes the necessity of resonant systems, whereas PC necessitates future prediction; reservoir computing (RC) incorporates both these attributes. In numerous works, the connectivity matrix within the reservoir is typically randomly generated~\cite{tanaka2022reservoir}. By judiciously adjusting the leaky rate and spectral radius, the reservoir can achieve one-step-ahead prediction in chaotic systems across various time scales~\cite{tanaka2022reservoir, manneschi2021exploiting}. A study by Gallicchio et al.~\cite{gallicchio2022euler} designed the connectivity matrix to adhere to an antisymmetric rule, yielding commendable performance on tasks involving long-term memory. Numerous studies have proposed alternatives to the random structure of reservoirs by devising models with regular graph configurations. For instance, one approach involves a delay line~\cite{rodan2010minimum}, where each node solely exchanges information with the previous and following nodes, and another model called the concentric reservoir, introduced in~\cite{bacciu2018concentric}, comprises interconnected delay lines forming a concentric structure. Additionally, the concept of a hierarchical architecture within ESNs, where each ESN connects to its preceding and following counterparts, has intrigued the reservoir computing community due to its ability to uncover higher-level features within external signals~\cite{gallicchio2017deep}.

\section{Methods}
\label{methods}

The main objective of the current model is to predict the occurrence of rhythmic beats.

Our innovation contains two steps. First, the new reservoir structure, which is introduced in \ref{section:structure}. During training it is combined with classical output weight training. Second, the new structure allows tuning the reservoir during the application phase. By means of our specifically designed synchronization loss function and the DS mechanism, adjustments are made to the reservoir's connection matrix to align the output predictions with our target. This ensures accurate matching between the predicted output and the desired target, as detailed in \ref{section:synchronization}.

\subsection{Reservoir weights}
\label{section:structure}

\begin{figure*}
    \centering
    \includegraphics[width=0.8\linewidth]{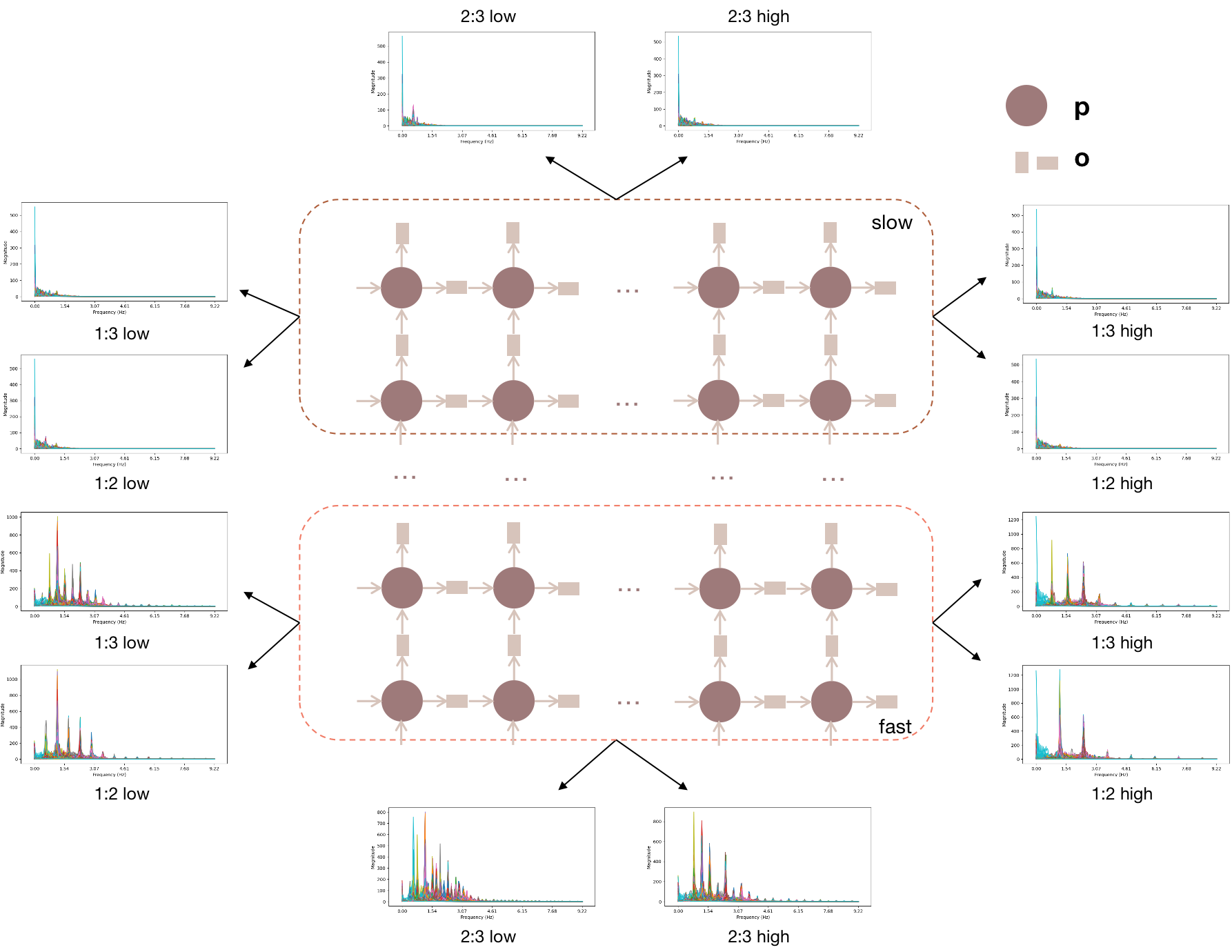}
    \caption{Overview of the model structure: This figure illustrates the components $p$, and $\mathbf{o}$ of the reservoir. Fast Fourier Transform spectra of select samples for both fast and slow segments are presented to highlight the resonances.}
    \label{fig:overview}
\end{figure*}

We start by introducing a classical reservoir computing (RC) method, the Echo State Network (ESN)~\cite{jaeger2007echo}. The ESN is a recurrent neural network. Its hidden states change according to the current input and the hidden states from the previous time step, which follows the equations: 
\begin{equation}
\begin{aligned}
    \mathbf{x}(t+\delta t) &= (1-\alpha)\mathbf{x}(t)+\alpha f(\mathbf{h}(t)), \\
    \mathbf{h}(t) &= \mathbf{W}_{in}\mathbf{s}(t)+\mathbf{W}\mathbf{x}(t) + \mathbf{W}_{bias},
\end{aligned}
\label{eq:reservoir}
\end{equation}
where $\mathbf{W}$ is a sparse matrix defining the connectivity of the network, $\mathbf{W}_{in}$ is the input weight, and $\mathbf{W}_{bias}$ is a bias matrix, and $\alpha$ is the leakage rate of the model. $f(\cdot)$ is a non-linear function, for which $tanh(\cdot)$ is used in this paper. 

Conventionally, the weight matrix of the reservoir $\mathbf{W}$, the input weight matrix $\mathbf{W}_{in}$, and the bias matrix $\mathbf{W}_{bias}$ are randomly generated. In this paper, $\mathbf{W}_{bias}$ represents noise, regenerated every time step. The weight matrix $\mathbf{W}$ is usually adapted to keep all of its eigenvalues inside the unit circle~\cite{manneschi2021exploiting} in the z-domain, thereby assuring stability of this dynamic system in the linear regime.

For the problem at hand, predicting the occurrence of rhythmic beats in a human-like way, the poles in the z-domain of the $\mathbf{W}$ matrix and thus the resonances in the random reservoir are not optimally placed: (1) they span a frequency range that does not match human capabilities; (2) many of them are too much damped. To overcome this problem, we propose a novel reservoir structure designed following physical principles. To simplify the tuning of $\mathbf{W}$, we design it based on a 2D Finite-Difference Time-Domain (2D-FDTD) computational approximation of the linearized Euler equations~\cite{botteldooren1995finite} for wave propagation in a medium with randomly generated properties. Starting from the wave equations Eq.~\ref{eq:wave_eq} where $c$ is the wave speed and $k$ is a damping (amplification if negative) factor, and $p$ and $\mathbf{o}$ are proportional to a pressure and velocity vector, the FDTD approximation generates a coupling matrix $\mathbf{A}$. A staggered grid, central differences, and an explicit time stepping are used. Stability is guaranteed by setting the courant number to 1.
\begin{equation}
\begin{aligned}
    \frac{\partial p}{\partial t} + c^2 \nabla \mathbf{o} &= 0 \\
    \frac{\partial \mathbf{o}}{\partial t} - k \mathbf{o} + \nabla p &= 0,
\end{aligned}
\label{eq:wave_eq}
\end{equation}

The weight matrix $\mathbf{W}$ of the reservoir is computed by:
\begin{equation}
\mathbf{W} = (\mathbf{A} - (1 - \alpha) \cdot \mathbf{I}) / \alpha,
\end{equation}
where $\mathbf{I}$ is the identity matrix. In this way the update equations of the reservoir Eq.~\ref{eq:reservoir} become very similar to the FDTD update equations. It implies very strong symmetry constraints on the $\mathbf{W}$ matrix. The two groups of unknowns could be interpreted as two types of neurons as in Fig.~\ref{fig:overview}: one is the primary neuron denoted as $p$, and the other is the intermediate neuron, labeled as $\mathbf{o}$. $p$, $\mathbf{o}$ are coupled locally and sparsely by $\mathbf{A}$. The local value of $c$ determines how strongly the $p$-neuron responds to inputs from surrounding $o$-neurons and together with the coupling to its neighbors this can result in local resonances, where the physics equivalent learns that small $c$ correspond to low-frequency resonances. By introducing a gradient in c on top of the random value, a slow (low-frequency resonances) and fast (high-frequency resonances) end of the reservoir can thus be realised as Fig.\ref{fig:overview}. The variable $k$ determines how much information is transferred between the $p$ neurons that the $o$-neuron connects. Increasing $k$ will result in more strongly damped resonances.

\subsection{Tuning for synchronization}
\label{section:synchronization}

In this study, we employ stochastic gradient descent (SGD) to minimize the Mean Squared Error (MSE) between the prediction $P$ and target $T$ signals during training that is based on a large number of rhythmic beats that could theoretically be encountered in music. Following training, the output layer can thus identify a suitable combination of correct oscillators, thereby providing an initial estimate of the target beat periodicity and timing.

To synchronize the prediction with the target beats more accurately, an adaptation phase is introduced. We divide the problem into two parts: `too early' or `too late', and assign corresponding errors for each case. In both scenarios, there is usually an overlap between the sound peaks corresponding to a beat in the prediction and target. Thus, the slope of the peaks is employed to calculate the error $I_{early}$ and $I_{late}$ of the prediction $P$ and target $T$ signals. 
If the prediction is descending while the target is ascending, we consider the prediction to be too early. Otherwise, if the prediction is ascending while the target is descending, the prediction is too late. We start accumulating $I_{early}$ and $I_{late}$ every time step from the beginning of a certain time interval $\text{update\_step}$, and reinitialize to 0 when the interval ends, as shown in Algorithm~\ref{alg:update_c}. 
To ensure proximity in amplitude between the target and prediction within the same time window, we first apply a moving average and a $softmax$ to normalize (eq. \ref{eq:norm_T}, \ref{eq:norm_P}) both the target and prediction values:
\begin{align}
T_{norm}(t) &= \frac{T(t) - T_{mean}}{T_{softmax}(t)}, \label{eq:norm_T}\\
P_{norm}(t) &= \frac{P(t) - P_{mean}}{P_{softmax}(t)},
\label{eq:norm_P}
\end{align}
where
\begin{align}
T_{softmax}(t) &= \ln (\int_{0}^{t} e^{T(t')} e^{\frac{t'-t}{\tau}}dt'), \\
P_{softmax}(t) &= \ln (\int_{0}^{t} e^{P(t')} e^{\frac{t'-t}{\tau}}dt').
\end{align}
where $\tau$ is an exponential averaging time constant spanning multiple beats.

By weighted comparison of $I_{early}$ and $I_{late}$, the decision is made to increase or decrease the speed-up factor $\delta_c$ by a fixed amount as shown in Algorithm~\ref{alg:update_c}. If the prediction is too early, we thus decrease all elements of $c$ proportionally to their value; if it is too late, we increase $c$ it. In this way, the whole reservoir slows down or speeds up.

Secondly, an attention-like DS mechanism is proposed to control the damping of the oscillations in the reservoir. The poles of the $W$ matrix are modified by identifying regions within the reservoir that are pivotal for accurately predicting beats and amplifying them by lowering $k$ in these regions. Simultaneously, the oscillations that make minor contributions are damped. To this end, each neuron within the reservoir is masked, generating masked outputs, and computed their MSE in comparison to the target in every time window. The neurons that resulted in the most significant MSE reduction when masked are considered the ones contributing the most to accurate prediction. Conversely, those neurons leading to the least reduction were considered to have the smallest contribution. We modulated the activity of these neurons by adjusting the parameter $k$ around their positions, either enhancing or diminishing their activity accordingly. As the change in $k$ remains fixed immediately after the target changes, this mechanism can be considered to focus attention on that area in the reservoir for some time period.

\begin{algorithm}[]
\caption{Calculate loss function to adapt $c$}
\label{alg:update_c}
\begin{algorithmic}[1]
\State Init: batch\_size, $\text{update\_step}$, \text{threshold\_sum}, \text{threshold}
\For{$i$ in batch\_size}
\State $\epsilon_{early}$ = 0, $\epsilon_{late}$ = 0, $\delta_{sum}$ = 0, $\delta_{c} = 0.02$
\For{$t$ \text{in} $\text{update\_step}$}
    \If{$T_{norm}(t) > \max(P_{norm}(t), 0)$}
        \If{$T_{norm}(t) - T_{norm}(t-1) > 0$} \If{$P_{norm}(t) - P_{norm}(t-1) < 0$}
            \State $I_\text{early}(t) = I_\text{early}(t-1) + 1$
        \EndIf
        \ElsIf{$T_{norm}(t) - T_{norm}(t-1) < 0$}
        \If{$P_{norm}(t) - P_{norm}(t-1) > 0$}
        \State $I_\text{late}(t) = I_\text{late}(t-1) + 1$
        \EndIf
        \EndIf
    \EndIf
    \State  $\epsilon_{early} = \epsilon_{early} +  \delta_{early} \cdot I_{\text{early}}(t)$
    \State  $\epsilon_{late} = \epsilon_{late} + \delta_{late} \cdot I_{\text{late}}(t)$
    \If{$\delta_{sum} < \text{threshold\_sum}$}
    \If{$\epsilon_{early} - \epsilon_{late} < \text{threshold}$}
        \State $\delta_{sum} = \delta_{sum} + \delta_{c}$
        \State $c *= 1 +\delta_{c}$
    \Else
        \State $\delta_{sum} = \delta_{sum} - \delta_{c}$
        \State $c *= 1 - \delta_{c}$
    \EndIf
    \Else
        \State No update.
    \EndIf
\EndFor
\EndFor
\end{algorithmic}
\end{algorithm}

\section{Experiments}

In this section, we conduct experiments on a benchmark dataset to evaluate whether our proposed approach reaches the objective of predicting the beat 200ms before it occurs. This time is roughly the time needed for a human to generate a motor action and would also allow an embodied system to act and e.g. hit a drum. We make a comprehensive analysis of our approach and compare it with other methods on the same task.

\subsection{Data}
In this study, we utilized data derived from uni-variable rhythmic beat signals for training and testing. Our training set comprised 1,000 generated samples characterized by varying beat frequencies ranging from 66 to 168 beats per minute. Additionally, we incorporated 25\% of non-rhythmic signals to further diversify the data set. Each sample's length is 30 seconds and the time step is 6 ms. Both the starting and ending points of the signals were randomly selected to maintain a diverse representation. 

To facilitate straightforward comparisons with other models, our test set consisted of 6 predefined samples with different frequencies, in other words, different inter-beat intervals.

\subsection{Experimental settings}

The size of the reservoir core is $n * n$, where $n = 40$ in this paper. This results in 1600 p-neurons and 3200 o-neurons. The matrix $c$ is initialized to 300, with a gradient decreasing from the fast end to the slow end at a rate of $-250/n$. Finally, the matrix is subtracted by a random matrix uniformly distributed between 0 and 0.8.
This design assured that the poles of $W$ correspond to frequencies matching the range of beat frequencies mentioned above. The $k$ is initialized to 0. Bias matrices are randomly initialized at every time step using a uniform distribution to introduce system noise.

The leakage rate, denoted as $\alpha$, is established as 0.03, computed as the reciprocal of $\tau$~\cite{yang2019task}, where $\tau = \frac{200}{6}$, reflecting our prediction of 200ms forward and downsampling by a factor of 6. The input weight is randomly initialized using a uniform distribution but only fed to the first row of 40 p-neurons. The output layer is trained to extract a linear combination from the information in the p-neurons. Thus, only the output layer undergoes training, accomplished through Stochastic Gradient Descent (SGD) with a fixed learning rate of 0.01. The loss function is the MSE between the prediction and the target shape.  

To enhance prediction precision, we use a synchronization loss mechanism to adapt the weight matrix~\ref{section:synchronization}. During the initial training phase, the loss function for synchronization remains inactive. However, once the pre-trained model is obtained, we introduce the loss function to synchronize predictions with target values at intervals $update\_step$ of 200 time steps. For adapting $c$, the $\mathrm{threshold}$ is initialized as 100. $\delta_c$ is the changing rate of $c$ and is set to 0.02, and $\delta_{sum}$ can not be larger than $\mathrm{threshold\_sum} = 10\%$. 
For adapting $k$, we selected 40 neurons that exhibited the most significant reduction in MSE and another 40 neurons with the least contribution. We altered the strength $k$ of the $\mathbf{o}$ located below and to the left of the $p$ by a fixed value of 0.002. When the neuron contributed to a decrease in MSE, we reduced $k$; conversely, if the neuron led to an increase in MSE, we increased $k$.

\subsection{Results}
We apply the proposed method to the fixed dataset and evaluate it on the test set. To validate our reservoir's ability to capture multiple frequencies and various combinations thereof, we input signals with two distinct frequencies into the reservoir. Subsequently, we analyze all frequencies within the range of $p$ in every example. The two input rhythmic signals with frequencies proportions of 1:2, 1:3, or 2:3, upon processing the entire sample by the model, we perform Fast Fourier Transform (FFT) analysis on all primary p-neurons, as depicted in Fig.\ref{fig:overview}. We use two different basic frequencies, a low one 1.2 Hz and a high one 2.4 Hz. The signal gradually attenuates as it propagates from the fast segment where it is injected to the slow segment within the reservoir. This decay can be observed in the spectra of the same sample within both fast and slow segments. The spectra also indicate that the reservoir responds to frequencies in the range of humanly plausible beats, suggesting that theoretically, the output layer can identify and predict frequencies accurately. In the 1:2 and 1:3 scenarios, the primary neurons within the reservoir not only capture the fundamental frequency but also exhibit peaks at half and one-third of the frequency. In the 2:3 scenario, the reservoir captures not only the fundamental frequency, half, and one-third of the frequency but also one-sixth of the frequency. As expected, the slow end of the reservoir only responds to these lower frequencies. 

For a more comprehensive study of the model's characteristics and performance, as well as comparative analysis with other models, we restricted the model usage to a single-input single-output setting in the following parts. Our evaluation method for the results involves assessing the differences between every peak in predictions from the test signal and the corresponding peak in the target signal, expressed as the time offset ratio. This ratio is computed as $(\mathrm{t_{target\_beat\_occurrence}} - \mathrm{t_{pred\_beat\_occurrence}})/\mathrm{target\_inter\_interval}$ here.

\begin{figure}
    \centering
    \includegraphics[width=0.95\linewidth]{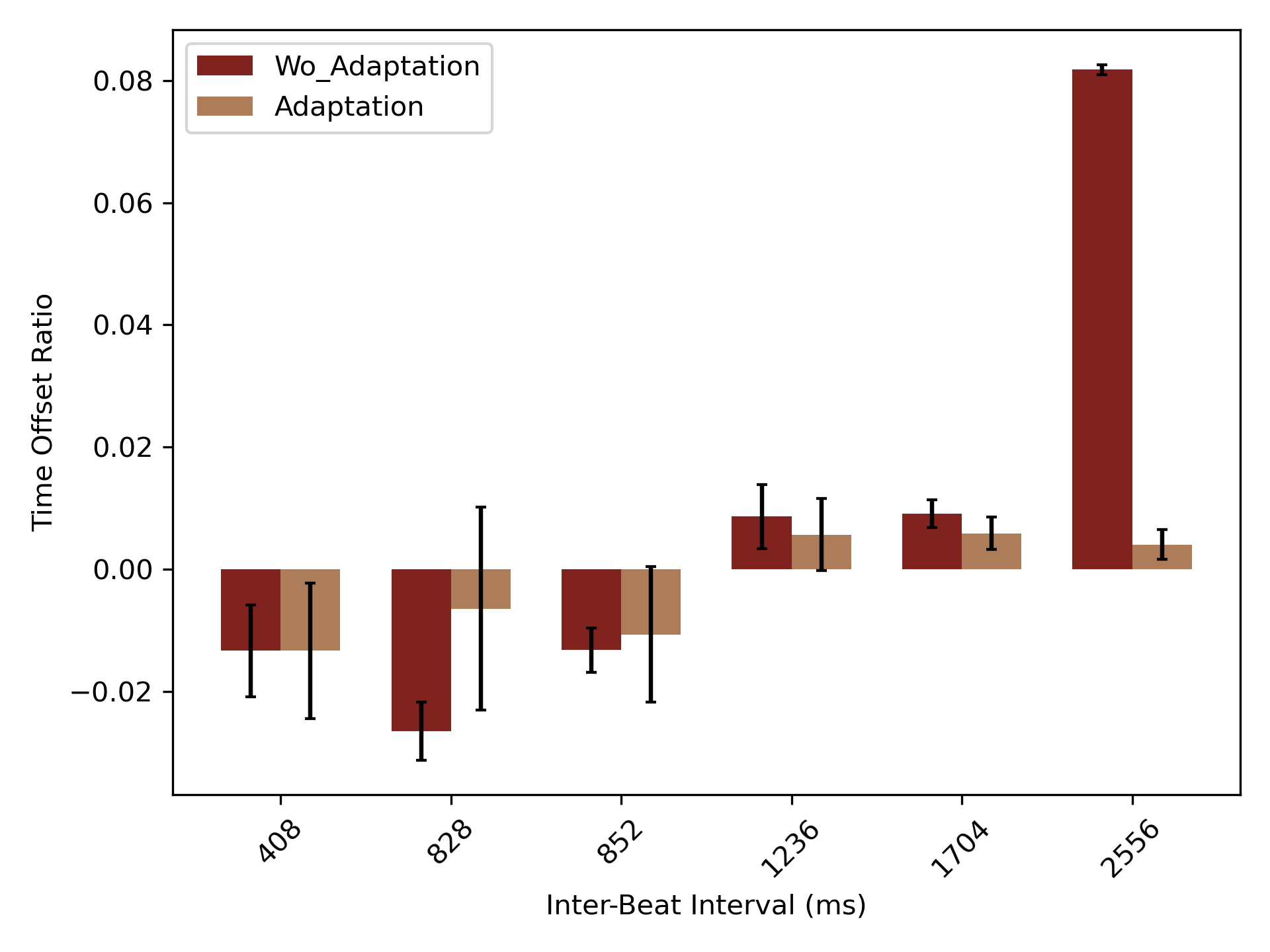}
    \caption{Comparing the time offset ratios pre- and post-adaptation through the assessment of mean and variance measurements.}
    \label{fig:result_adapt}
\end{figure}

The examples of comparing the prediction and the target rhythmic signals' mean and variance of time offset ratio are shown in Fig.~\ref{fig:result_adapt}. We compared six sets of different frequencies to analyze the time offset ratio between the model's outputs before and after adjustments. The figure illustrates that in the absence of adaptation, the rhythm's offset remains small, with a relative proportion to the inter-beat interval not exceeding 4\%. This observation indicates the reservoir's ability to accurately predict periodic signals. By employing our proposed synchronization algorithm to adjust parameters $c$ and $k$, we alleviate instances where the original model's output leads or lags compared to the rhythm in the target. Post-adaptation, the mean offset ratio diminishes below 1\%, accompanied by a decrease in variance as depicted in the figure. This outcome suggests that our synchronization algorithm reliably and consistently adjusts the output. Note that there is a trend from anticipation to lagging as the inter-beat interval increases. 

\begin{figure}
    \centering
    \includegraphics[width=\linewidth]{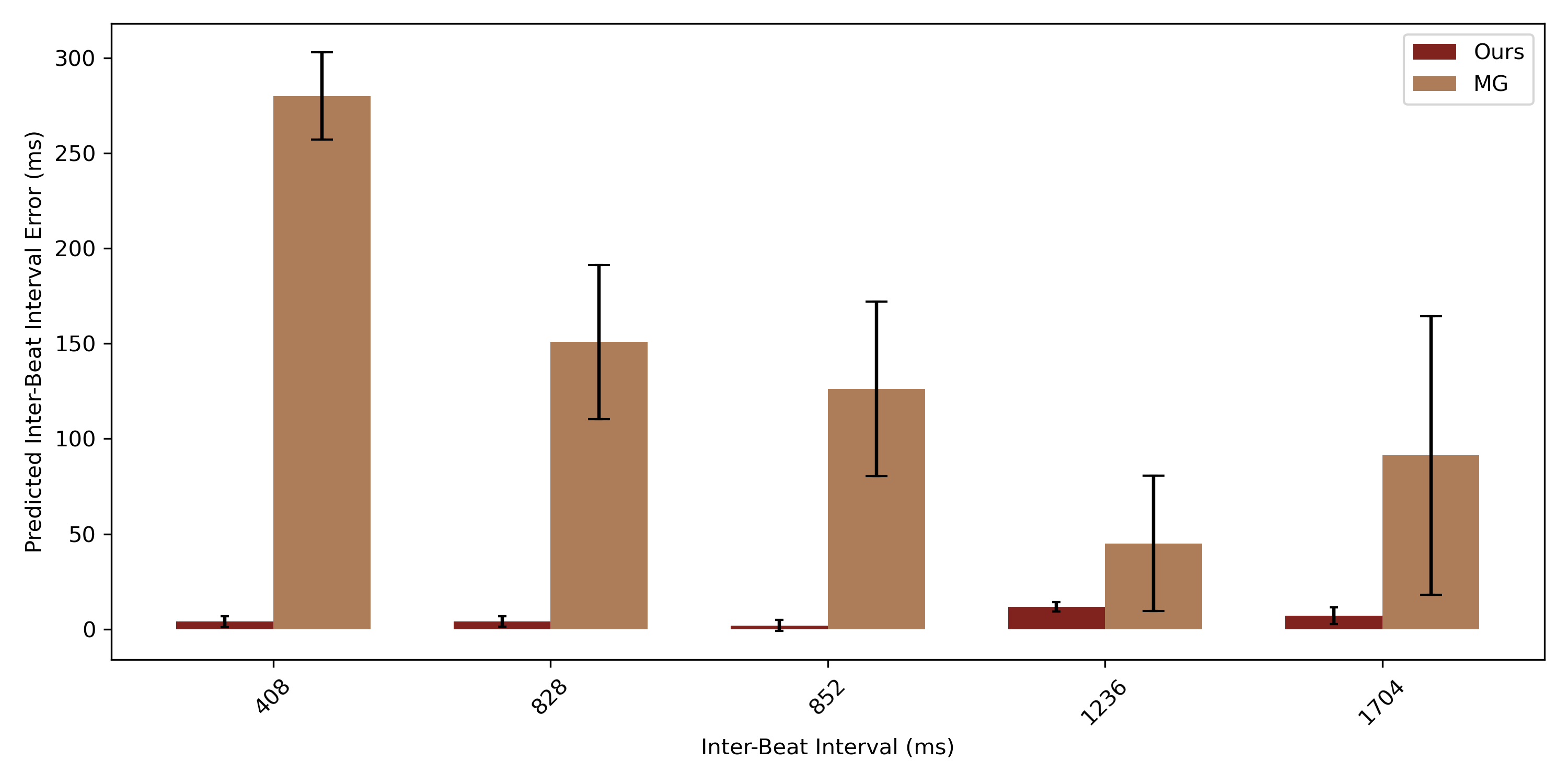}
    \caption{Evaluating Synchronized Time Intervals: Mean error and variance calculations are performed on the discrepancies between intervals generated by the motor generator and the target intervals, alongside predictions made by our model.}
    \label{fig:time_interval}
\end{figure}

\subsection{Comparison to other models}

To validate our proposed approach, we compare it with two other methods: the most basic randomly generated reservoir, and a motor generator from ~\cite{egger2020neural}.

Commonly, the weight matrix of the reservoir is randomly generated. Here, we randomly generated a sparse connection matrix, while keeping the input layer and bias matrix unchanged. After training using the same method, the rhythm generated by the random reservoir follows the target with 200ms delay, which means the model is not predicting.

The motor generator is constructed by two differential equations. It primarily operates by learning rhythmic patterns with fixed intervals. Occasionally, phase-related issues may arise, where the target exhibits two peaks, while the synchronized result may display three. Therefore, using the interval length between peaks makes it easier to assess synchronization performance. The comparison of the synchronized time interval error is shown in Fig.~\ref{fig:time_interval}. The actual time intervals between peaks are calculated. This figure shows the mean and variance of this error in time intervals between the predicted outputs generated by the method proposed in this paper, the motor generator, and the actual target. As the motor can not follow the beat for the 2556 ms inter-beat interval, it is not listed in the figure. In other cases, the reservoir can capture the time interval better, with smaller mean errors. Our model exhibits smaller inter-beat interval errors, indicating that it is less prone to phase misalignment.

\subsection{Ablation Study}

\begin{figure}
    \centering
    \includegraphics[width=\linewidth]{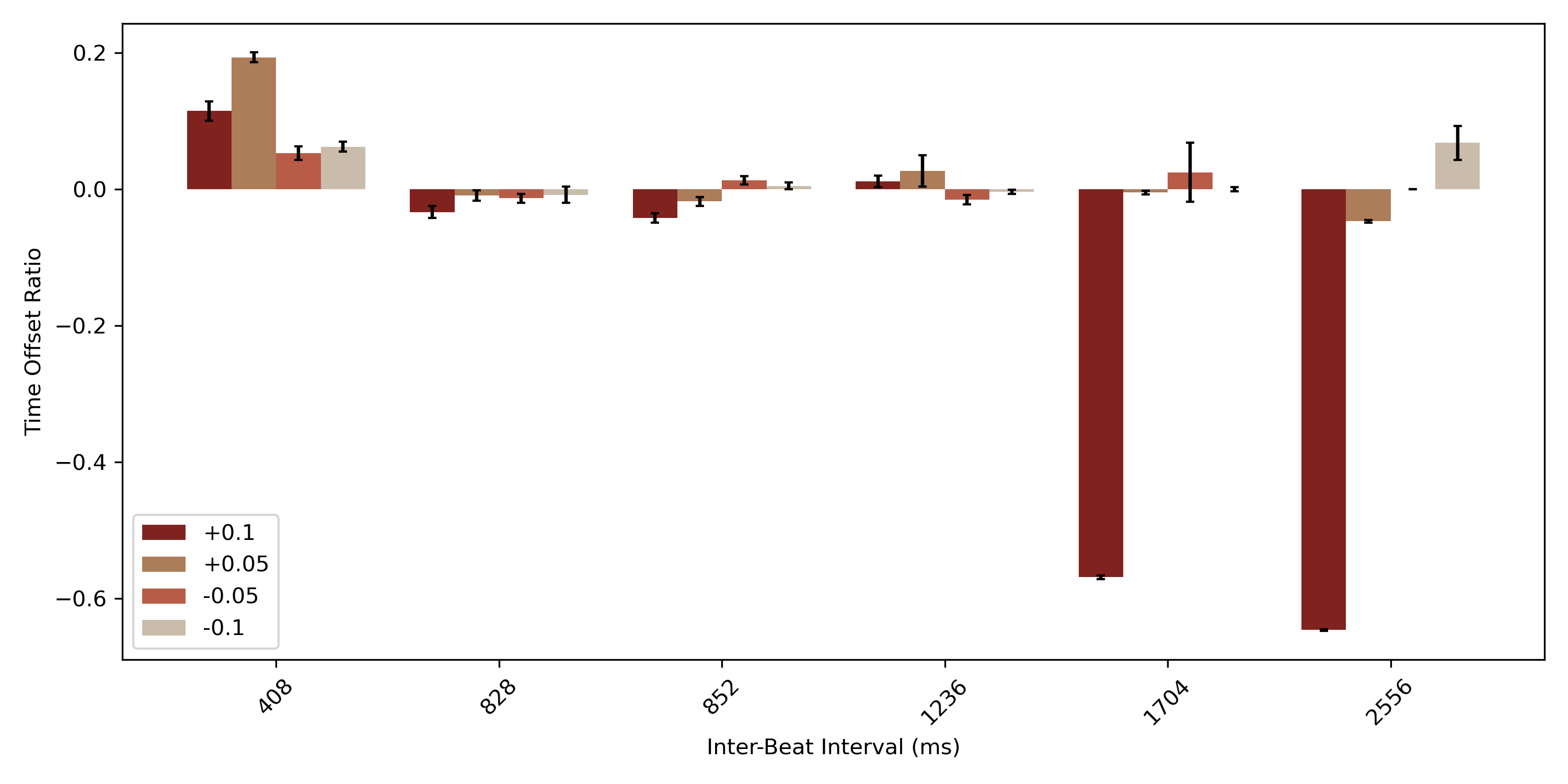}
    \caption{An illustration showcasing the effectiveness of adapting matrix c is presented. Various constant values of $\delta_c$ are employed for the adaptation process, and the resulting mean error and variance between predictions and targets are computed to highlight the performance of the adaptation.}
    \label{fig:adapt_c}
\end{figure}

To explore how changing the reservoir speed matrix $c$, affects the time offset ratio $\mathrm{(t_{adapted\_beat\_occurrence}} - \mathrm{t_{prediction\_beat\_occurrence}})/\mathrm{prediction\_inter\_interval}$, the pre-trained model without adaptation is applied to the test set after adapting $c$ with a few fixed $\delta_{c}$ = \{-0.1, -0.05, +0.05, +0.1\}. We measured the time offset ratio before and after adaptation for the four values of $\delta_{c}$ and calculated the mean and variance. The results are shown in the Fig.~\ref{fig:adapt_c}. From the figure, it is evident that in most cases increasing $c$ shifts the peak's occurrence towards an earlier time, while decreasing $c$ shifts it towards the opposite way.

At this point, the effect of introducing the DS mechanism to guide the model's focus towards the desired oscillation is investigated. 
The proposed attention mechanism is designed to adjust the damping matrix $k$, which represents the loss during the transition of information between p-neurons and therefore also from the fast area of the reservoir to the slow area. We remind the reader that we adjust $k$ to amplify the oscillations within the reservoir that contribute more effectively to accurate prediction, while concurrently dampening those oscillations that contribute less and have negative effects. 

\begin{figure}[!ht]
  \centering
  \begin{minipage}[b]{\linewidth}
      \begin{minipage}{\linewidth}
        \centering
        \includegraphics[width=0.48\linewidth, trim=35 23 40 40, clip]{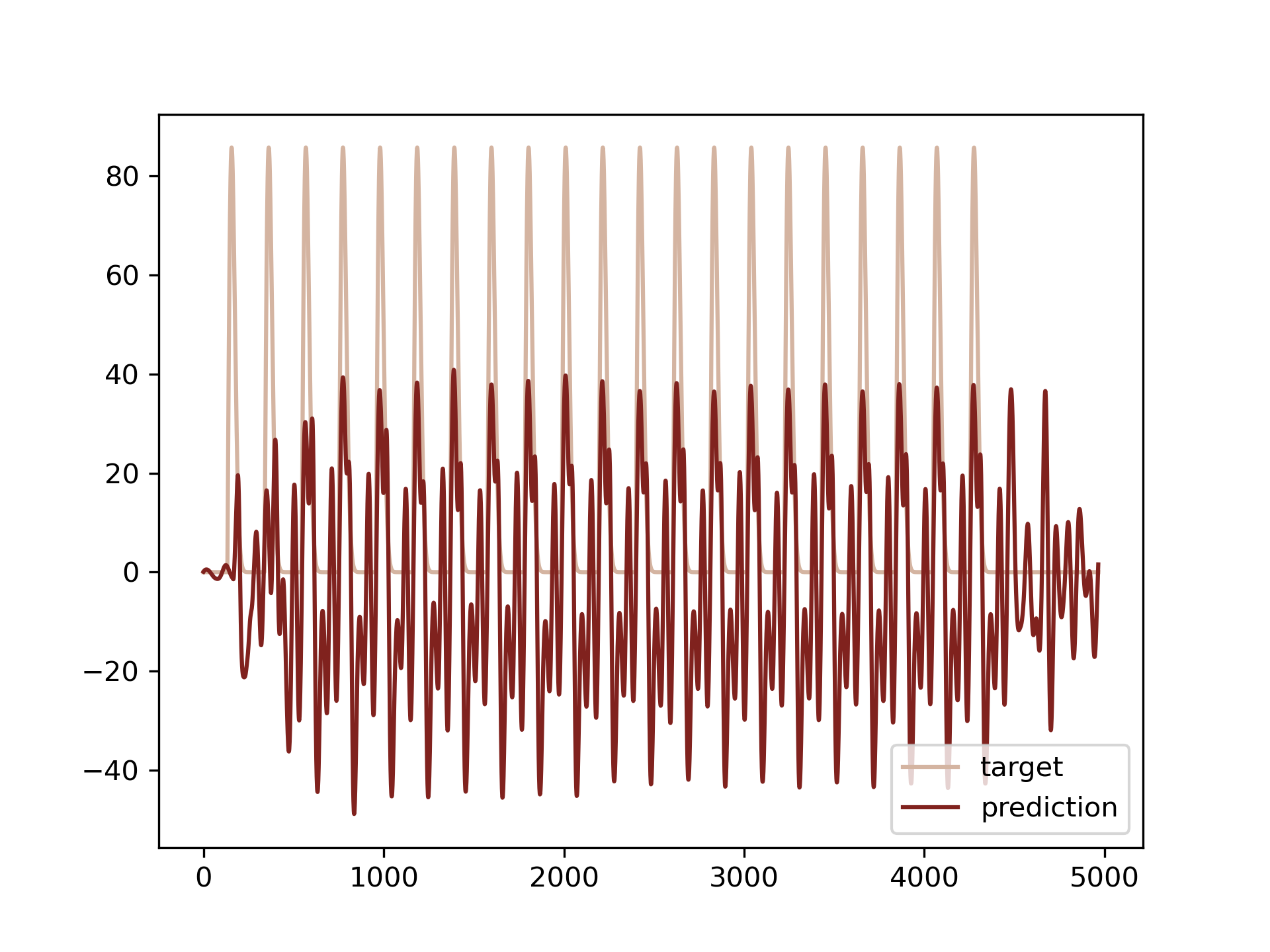}\vspace{8pt}
        \includegraphics[width=0.48\linewidth, trim=35 23 40 40, clip]{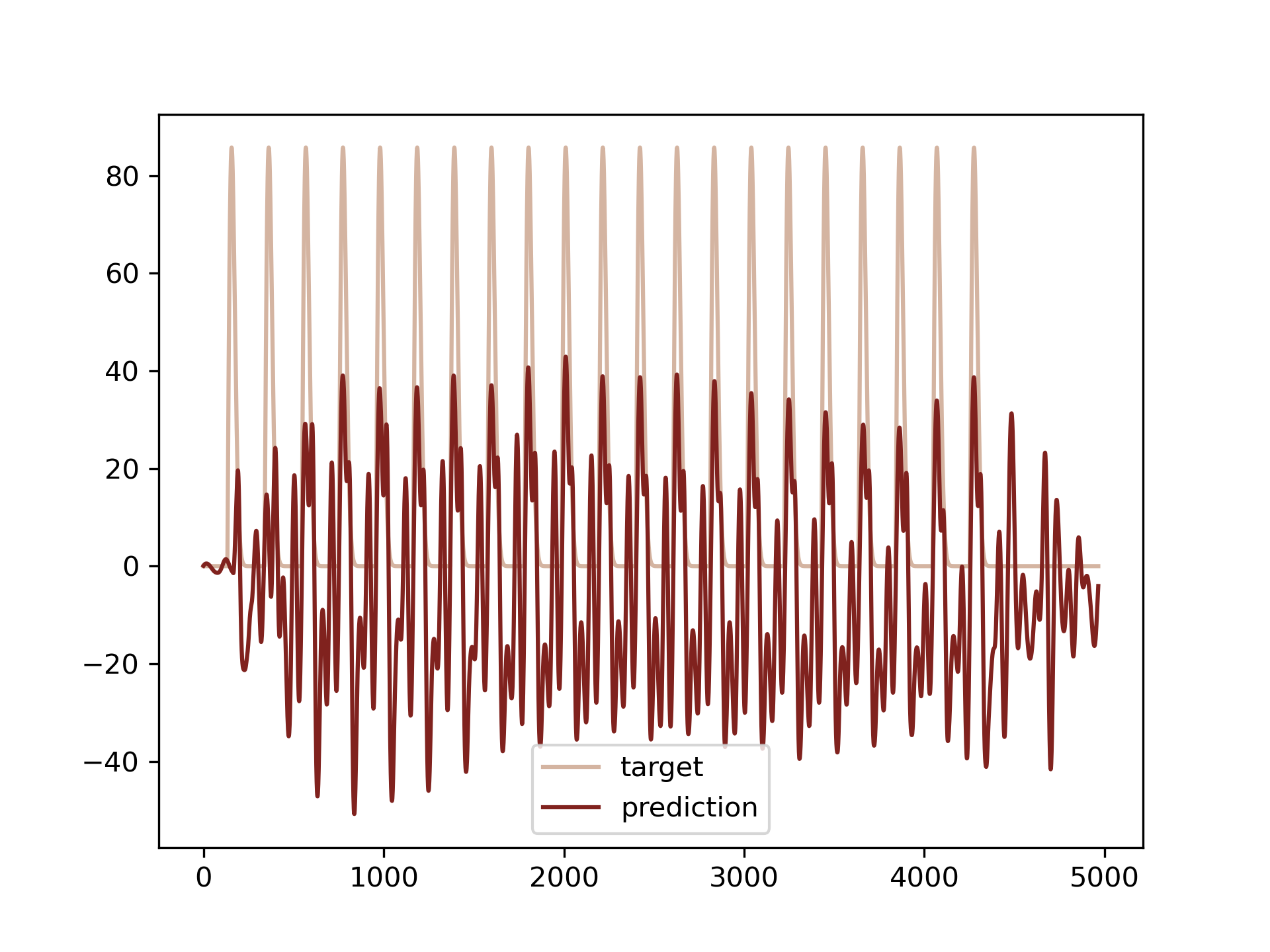}
      \end{minipage}
        \captionsetup[subfigure]{labelformat=empty} 
        \subcaption{(a) The comparative panel between predictions and targets before and after adjusting parameter $k$.}  
    \hfill
      \begin{minipage}[b]{\linewidth}
        \centering
        \includegraphics[width=0.48\linewidth, trim=40 60 40 40, clip]{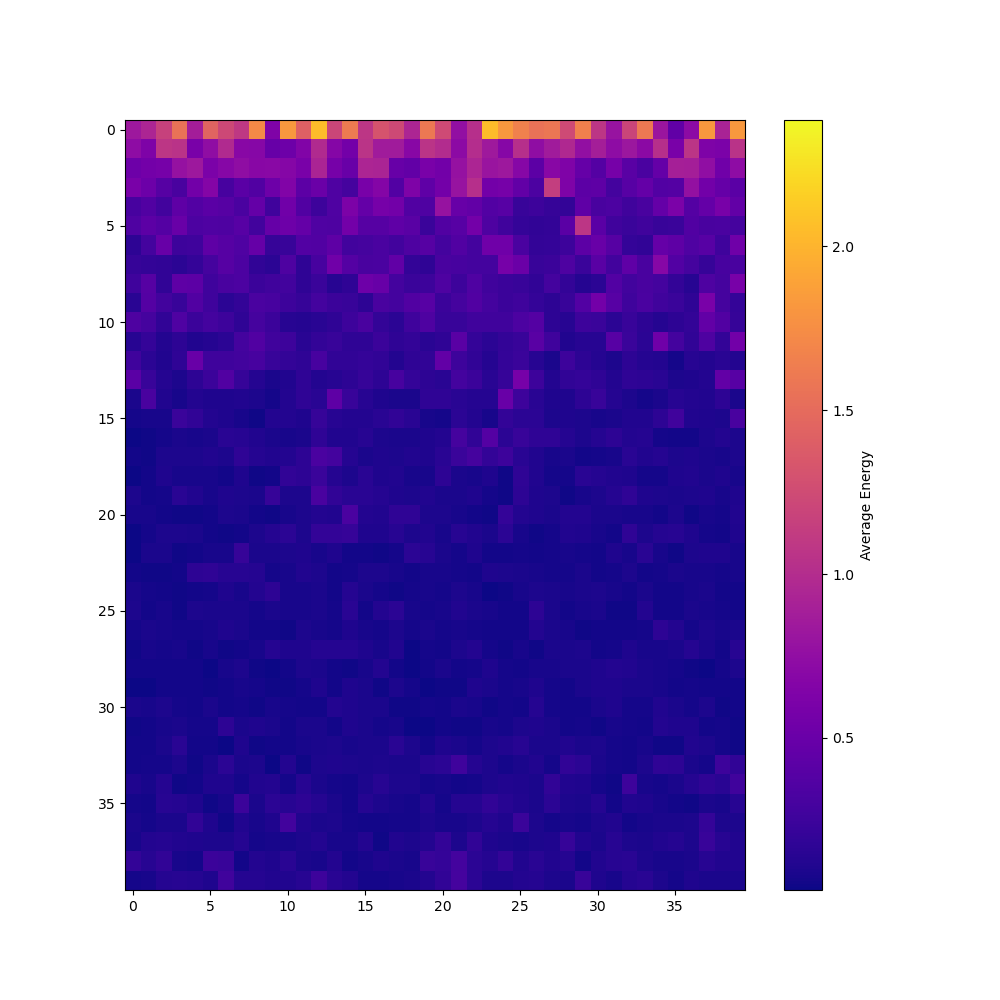}\vspace{8pt}
        \includegraphics[width=0.48\linewidth, trim=40 60 40 40, clip]{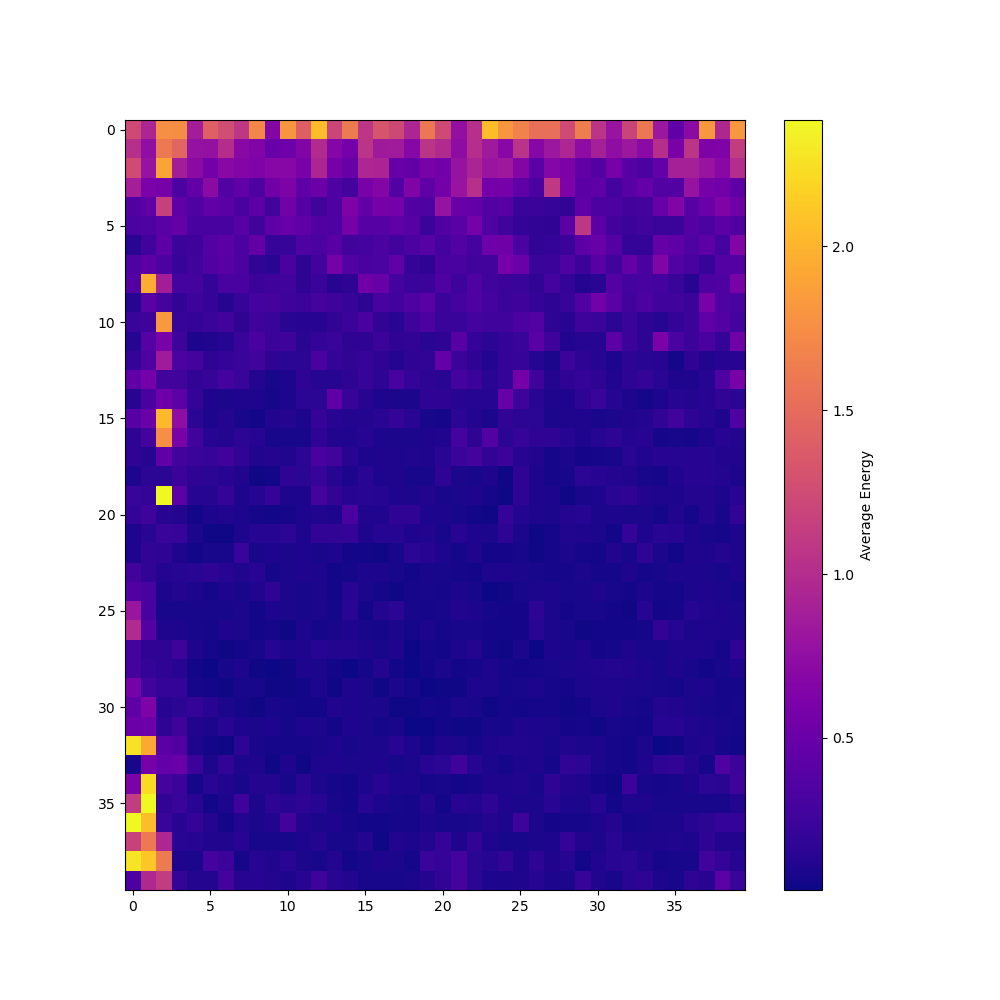}
      \end{minipage}
        \captionsetup[subfigure]{labelformat=empty} 
        \subcaption{(b) The panel compares the average signal strength heat map of all primary neurons before and after the adjustment of parameter $k$ after going through this example.}  
  \end{minipage}
  \vfill
  \caption{The influence of adjusting parameter $k$ is evident in the figures. The image on the left illustrates the outcomes before this adjustment, while the one on the right depicts the post-adjustment effects. Panel (a) visualizes the alterations in predictions both before and after the adaptation of $k$, while panel (b) presents the average activation of primary neurons, obtained after processing the entire sample.}
  \label{fig:adapt_k}
\end{figure}

In Fig.\ref{fig:adapt_k}, an illustration is presented. In Panel (a) of this figure, it is evident that during the update of parameter $k$, the amplitude of undesired peaks relative to the desired peaks gradually diminishes, eventually reducing completely below zero. This demonstrates the efficacy of our adaptation of $k$. In the comparative analysis shown in Panel (b), the signal strength of primary p-neurons is shown, indicating shifts in active areas, consequently altering the model's output.

\section{Conclusion}

This paper introduces a novel reservoir model structure tailored for predicting rhythmic time series that could be encountered in music. To this end, the reservoir weights are based on a numerical approximation of wave equations. This allows (1) to more easily design the reservoir to model strong resonances at frequencies that match human beat following capabilities, notably boosting performance in this domain; (2) spatially separate the regions that respond to slow and fast oscillations thereby giving direct access to these areas to construct the desired output; (3) to quickly tune overall behavior of the reservoir during the prediction phase. Through adaptive adjustment of propagation speed within the reservoir, we achieved enhanced synchronization, thereby yielding more precise predictions. Furthermore, an attention mechanism has been integrated that explores the option to adapt the decay rate, influencing reservoir oscillations to facilitate easier detection of beats in predictions. Our experiments, conducted on a consistent test set, showcased the superior performance of our algorithm over other human-like beat prediction models.

\section{Acknowledgments}
The authors gratefully acknowledge the financial support for this work provided by the BOF grant (BOF/24J/2021/246) and the WithMe FWO grant (3G043020). This research was also partially supported by the Flemish AI Research Programme.

\bibliographystyle{IEEEtran}
\bibliography{ref}

\vspace{12pt}
\color{red}

\end{document}